# Toward Psycho-robots

Andrei Yu. Khrennikov

Center for Mathematical Modeling in Physics and Cognitive Sciences,
University of Växjö, S-35195, Sweden
Email: Andrei.Khrennikov@msi.vxu.se

**Abstract:** We try to perform geometrization of psychology by representing mental states, "ideas," ideas by points of a metric space—mental space. Evolution of ideas is described by dynamical systems in metric mental space. We apply the mental space approach for modeling of flows of unconscious and conscious information in the human brain. In a series of models, Models 1-4, we consider cognitive systems with increasing complexity of psychological behavior determined by structure of flows of ideas. Since our models are in fact models of the AI-type, one immediately recognizes that they can be used for creation of AI-systems, which we call psycho-robots, exhibiting important elements of human psyche.

Creation of such psycho-robots may be useful improvement of **domestic robots.** At the moment domestic robots are merely simple working devices (e.g. vacuum cleaners or lawn mowers) . However, in future one can expect demand in systems which be able not only perform simple work tasks, but would have elements of human self-developing psyche. Such AI-psyche could play an important role both in relations between psycho-robots and their owners as well as between psycho-robots. Since the presence of a huge numbers of psycho-complexes is an essential characteristic of human psychology, it would be interesting to model them in the AI-framework.

## 1. Introduction

The aim of this paper is to present a mental AI-model based on a geometric representation of mental processes. This model can be considered as the first step in coming AI-formalization of foundations of psychoanalytic research. [1] Mathematical foundations for the present AI-model were developed in a series of works Khrennikov, 1997, 1998a,b, 1999a,b, 2000a,b, 2002a and Albeverio et al., 1999, and Dubischar et al, 1999. Unfortunately, the high level of mathematical presentation in these works makes them non readable for people working in AI, computer science, psychology. In the present article we would like to present the main distinguishing AI-features of our model without using the formal mathematical apparatus. Another important difference of this paper from mentioned works is that now we do not try to specify the set-theoretic and topological structure of mental space. In previous works we developed one special mathematical model of mental space given by hierarchical trees (so called utrametric spaces), see Khrennikov, 2004a, b, for neurophysiological basis for such spaces. Although such encoding of hierarchy into space topology is very promising (especially by taking into account the role of hierarchical structures in psychology), we find possible to proceed in modeling of flows of conscious/unconscious mind in the most general framework of arbitrary metric mental space. However, from the very beginning we emphasize that we could not exclude that practical creation of AI-psyche would be based on the hierarchical encoding of information by using trees equipped with ultrametric distance.

Our basic idea is to repeat in psychology and cognitive sciences the program of **geometrization** which has been performed in physics. And we hope that through such geometrization we shall be able to represent some elements of human psyche in the AI-framework. We recall that in physics the starting point of the mathematical formalization was creation of an adequate mathematical model of physical space. It was not so easy task. It took about three hundred years. However, finally, physicists got a well established model of space -- infinitely divisible real continuum. Physical systems where embedded in this space. Evolution of a physical system was represented by dynamical system (continuous—differential equation or discrete—itterations of some map from physical space into itself) . The basic dynamical law—the second Newton law—was given in a simple differential form. We would like to do the same with mind: a) to introduce mental space—"space of ideas" ; b) to consider dynamics in mental space—flows of ideas.

After performing such a nontrivial task, we apply our approach to modeling of Freud's psychoanalysis. Here we propose a mental AI-model describing flows of mind in the unconscious, the subconsciousness, and the consciousness as well as mental flows between these domains. The main attention will be paid to dynamics in the unconscious and the subconsciousness and their feedback coupling. Our model does not say so much about consciousness. Moreover, we are not sure that a mental AI-model could be applied to the problem of consciousness at all. Our model describes (and can be even used for a mathematical simulation) of such basic features of psychoanalysis as repression of forbidden wishes, desires and impulses (coming to the subconsciousness from the unconscious and going to the consciousness), complexes and corresponding symptoms.

In a series of Models 1-4, we consider AI-modeling of cognitive systems with increasing complexity of psychological behavior determined by the structure of flows of ideas. One immediately recognizes that our models can be used for creation of AI-systems, which we call **psycho-robots**, exhibiting important elements of human psyche. Creation of such psycho-robots may be useful improvement of **domestic robots.** At the moment domestic robots are merely simple working devices (e.g. vacuum cleaners or lawn mowers) . However, in future one can expect demand in systems which be able not only perform simple work tasks, but would have elements of human self-developing psyche.[2] Such AI-psyche could play an important role both in relations between psycho-robots and their owners as well as between psycho-robots. Since the presence of a huge numbers of psycho-complexes (results of repression of forbidden desires) is an essential characteristic of human psychology, it would be interesting to model them in the AI-framework.

---

[1] In spite of huge diversity of viewpoints on Freud´s psychoanalysis and its connections with neurophysiology, see, e.g., Macmillan, 1997, Gay, 1988, Young-Bruehl, 1998 as well as Solms and Turnbull, 2003, Green, 2003, Stein et al., 2006, Solms, M., 2002, 2006a, 2006b, for debates, the ideas of Sigmund Freud are still important sources of inspiration and not only in psychology, but also in cognitive sciences and even neurophysiology, neuro informatics and cybernetics as well as mental informatics and cybernetics.

[2] People would obtain not simple service-devices, but a kind of robots-relatives. A new possibility to build close psychological relations with a domestic robot could attract customers. Our key point is that robot-psyche should be really humanoid-like. Hence, such a robot should evolve all our habits which would induce with necessity main human psychological problems (including even psychopathic behavior), see our Model 4.

Our approach can be considered as extension of the artificial intelligence approach, Chomsky, 1963, Churchland and Sejnovski, 1992, to simulation of psychological behavior, cf. Boden, 1996, 1998, 2006. Especially close relation can be found with models of **AI-life**, see Langton et al., 1992 (and especially the article of Langton, 1992) , Yaeger, 1994, Collings and Jefferson, 1992. We extend modeling of AI-life to psychological processes. On the basis of the presented models, we can create AI-societies of psycho-robots interacting with real people and observe evolution of psyche of psycho-robots (and even people interacting with them). We also mention development of theory of **Animats**, see e.g. Meyer and Guillot, 1994 and Donnart and Meyer, 1996. By similarity with Animats we call our psycho-robots: **Psychots.**

Finally, as a motivation of our activity, we cite Herbert Simon: "AI can have two purposes. One is to use the power of computers to argument human thinking, just as we use motors to argument human or horse power. Robots and expert systems are major branches of this. The other is to use a computer's AI to understand how humans think, In a humanoid way…. You are using AI to understand the human mind." Our aim is precisely to understand human mind and psychology via AI-modeling.

## 2. Metric spaces

The notion of a **metric space** is used in many applications for describing distances between objects. There is given a set of objects of any sort. They are called points. There is defined a distance (metric) between any two points which is nonnegative and it has the following properties: 1). **separation:** the distance between two points equals to zero if and only if these points coincide; 2) **symmetry:** the distance between two points does not depend on order in which points are taken.; 3) **triangle inequality:** take three points and consider the corresponding triangle; each side of this triangle is less than or equal to the sum of two other sides.

The main examples of metric spaces which are used in physics are Euclidean spaces and their generalizations. However, as we have seen in Khrennikov, 1997, 1998a,b, 1999a,b, 2000a,b, 2002a and Albeverio et al., 1999, and Dubischar et al, 1999., another class of metric spaces might be essentially more adequate for applications to psychology and cognitive sciences, so called **ultrametric spaces** (in mathematical literature they are also called non-Archimedean spaces, Khrennikov, 1997). Those spaces have geometries which differ crucially from geometries of physical spaces. However, we are not able to go into detail in the present communication.

## 3. Dynamical thinking in mental space

We shall use the following mathematical model for mental space: **The set of mental states—"ideas" -- has the structure of metric space.**

Dynamical thinking, **evolution of a mental state**, is performed via the following procedure: a) an initial mental state (e.g. an external sensory input) is sent to the unconscious domain; b) it is iterated by some dynamical system[3] which is given by a map from the mental metric space into itself; c) if iterations converge (with respect to the mental space metric) to an attractor [4], then this attractor is communicated to the subconsciousness; this is the solution of the initial problem. [5] In the simplest model, see Model 1 in section 4.1, this attractor is sent directly to the consciousness.

Thus in our model unconscious functioning of the brain is **not based on the rule of reason.** The unconsciousness is a collection of dynamical systems (thinking processors) which produce new mental states practically automatically. The consciousness only uses and control results (attractors in spaces of ideas) of functioning of unconscious processors.

---

[3] The description of functioning of the human brain by dynamical systems (feedback processes) is a well established approach. The main difference between our approach and the conventional dynamical approach to cognition (see Ashby, 1952, van Gelder and Port, 1995, van Gelder, 1995, Strogatz, 1994, Eliasmith, 1996) is that in the conventional dynamical approach dynamical systems work in the real physical space of electric potentials and in our approach dynamical systems work in the mental space. Thus the conventional dynamical approach is about neuro processing of information, but our approach is about purely mental flows of information.

[4] We recall that a point of a metric space is called **attractor** of some dynamical system, if there exists a ball with the center at this point such that starting with any point of this ball as the initial condition iterations of the dynamical system will approach the attractor-point.

[5] Depending on initial conditions and the dynamical law it may occur that iterations starting with some initial mental state will not approach any attractor. For example, starting with some state a dynamical system may perform the cyclic behavior in the process of thinking. In such a case a cognitive system would not find the solution of a problem under consideration.

# 4. Transformation of unconscious mental flows into conscious flows

We represent a few mathematical models of the information architecture of conscious systems cf., e.g., Fodor and Pylyshyn, 1988, Edelman, 1989, Voronkov, 2002a. We start with a quite simple model (Model 1). This model will be developed to more complex models which describe some essential features of human cognitive behavior. The following sequence of cognitive models is related to the process of evolution of the mental architecture of cognitive systems.

## 4.1. Model 1: attractors

A). The brain of a cognitive system is split into three domains: **conscious, subconscious, and unconscious.**
B). There are two control centers, namely, **a subconscious control center** SCC and **an unconscious control center** UC.
C). The main part of the unconscious domain is a **processing domain.** Dynamical thinking processors are located in this domain. In our mathematical models such processors are represented by maps from mental space into itself.

In the simplest case the outputs of one group of thinking processors are always sent to the unconscious control center UC and the outputs of another group are always sent to the subconscious control center SCC.[6] The brain of such a cognitive system works in the following way, see section 3: a). external information (e.g., a sensor stimulus) is transformed by SCC into some initial idea-problem; the SCC sends this idea to a thinking processor which is located in the processing domain; b). Starting with this initial idea, the processor produces via iterations an idea-attractor.

We consider two possibilities:

c1). If the thinking processor under consideration is one of processors with the UC-output, then the idea-attractor is transmitted to the control center UC. This center sends it either as an initial idea to the processing domain or to an unconscious performance.
c11). In the first case some processor (it can have either conscious or unconscious output) performs iterations starting with this idea and it produces a new idea-attractor.
c12). In the second case there is produced some unconscious reaction.

c2). If the thinking processor under consideration is one of processors with the SCC-output, then the idea-attractor is transmitted directly to the control center SCC. This center sends it either again to the processing domain (as an initial idea) or to physical or mental performance (speech, writing), or to memory. Those performances can be conscious as well as unconscious. In the first case the idea-attractor should be transmitted by SCC to the conscious domain.

In this primitive model there is no additional analysis of the idea-attractor which was produced in the unconscious domain. Each attractor is recognized by the control center SCC as the solution of an initial problem, compared with models 2-4. Those attractors are wishes, desires and impulses produced by the unconsciousness. Moreover, it is natural to assume that some group of thinking processors have their outputs only inside the processing domain. Thus they do not send outputs to the control centers. An idea-attractor produced by such a processor is transmitted neither to SCC nor to UC. The idea-attractor is directly used as the initial condition by some processor.

---

[6] Information produced by processors with UC-outputs cannot be directly used in the subconscious or conscious domains. This information circulates in the unconscious domain. Information produced by processors with SCC-outputs can be directly used in the subconscious domain.

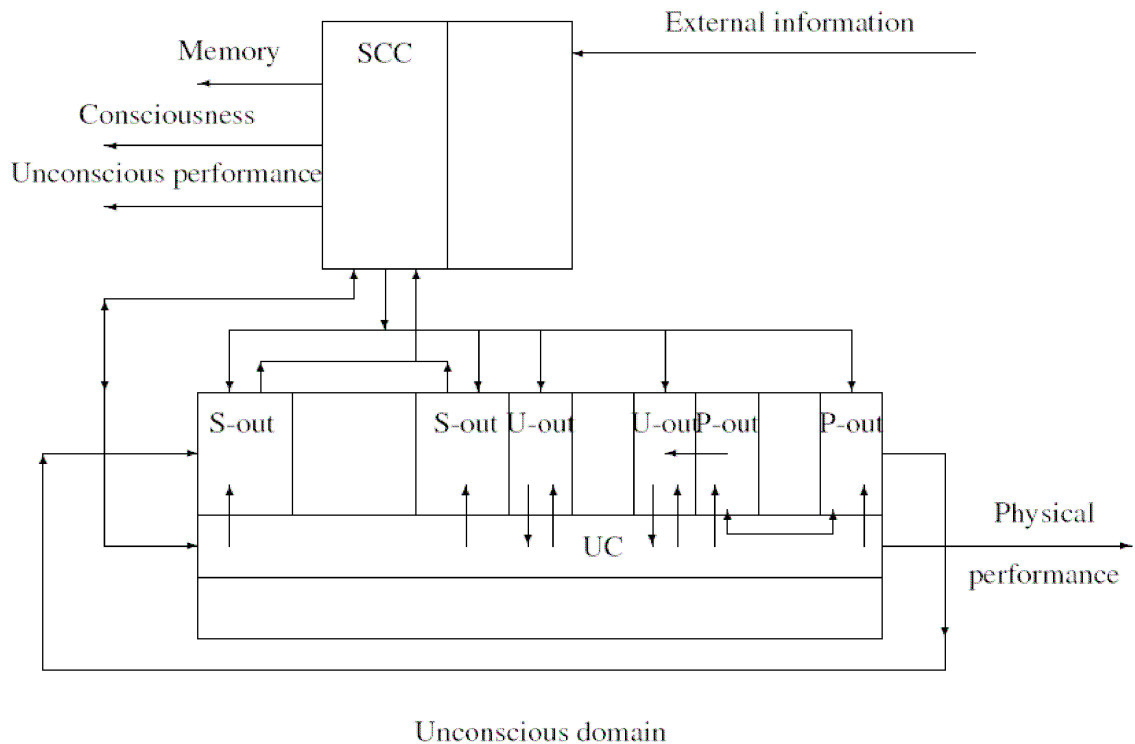

Figure 1: Model 1 of subconscious/unconscious functioning

**Comment on Figure 1:** Besides the unconscious control center UC and the processing domain, the unconscious domain contains some other structures (empty boxes of this picture). These additional structures (in the subconscious as well as unconscious domains) will be introduced in more complex models. We shall also describe the character of connections between SCC and UC.}

### 4.2. Model 2: measure of interest

One of the possibilities to improve functioning of a cognitive system is to create a queue of ideas waiting for realizations. Thus it is natural to assume that the subconscious domain contains some **collector** in that all "waiting ideas" are gathered.

Ideas in the collector must be ordered for successive realizations. The same order structure can be used to delete some ideas if the collector is complete. Thus all ideas-attractors must be classified.

Each idea-attractor obtains some quantitative characteristic that gives a **measure of interest** to this idea. We may assume that this quantity takes values in segment [k, 1], where k is a nonnegative real number (it depends on the range of values of the distance between mental points, see later on the concrete construction for the measure of interst).
If the measure of interest for some idea equals 1, then such an idea is extremely interesting for the cognitive system under consideration. If the measure of interest for some idea equals k, then the cognitive system is not at all interested in such an idea.

We assume that there exists a threshold of the minimal interest for realization -- **realization threshold.** If the measure of interest to some idea is less than the realization threshold, then the control center SCC directly deletes this idea, despite the fact that the idea was produced in the unconscious domain as the solution of some problem. If the measure of interest for some idea is larger than the realization threshold, then SCC sends this idea to the collector of ideas waiting for realization.

Any cognitive system lives in the continuously changed environment. It could not be concentrated on realization of only old ideas-attractors even if they are interesting. Realizations of new ideas which are related to the present instant of time t can be more important. The time-factor must be taken into account. Thus the measure of interest to any idea should depend on time and it should decrease with time. The speed of decreasing of the measure of interest can depend on the idea. Finally, if the measure of interest becomes less than the realization threshold such an idea-attractor is deleted from the collector.

It is natural to assume the presence of a **preserving threshold.** If an idea has an extremely high value of interest— which is larger than the preserving threshold, then such an idea must be realized in any case. In our model we postulate that for such an idea the measure of interest is not changed with time. We now describe one of the possible models for finding the value of interest for ideas-attractors. It is based on the fundamental assumption that the brain is able to measure the distance between ideas. The subconscious domain of the brain of a cognitive system contains a **database of ideas which are interesting for this system**. The interest-database is continuously created on the basis of mental experiences. It is the cornerstone of Ego (in coming models Ego will be essentially extended). The subconscious domain contains a special block, comparator, that measures the distance between two ideas, and the distance between an idea and the set of interesting ideas. At the present level of development of neurophysiology we cannot specify mental distance. Moreover, neural realization of mental distance may depend on a cognitive system or class of cognitive systems. However, the **hierarchic structure of the process of thinking** gives some reasons to suppose that functioning of a brain might be based on neuronal trees which induce ultrametric mental space, see Khrennikov 2002a, 2004a. Our present considerations are presented for an arbitrary metric. It is only important that the brain is able to measure the distance between ideas and between an idea and a collection of ideas.

We recall that the distance between a point (in our case an idea) and a finite set (in our case the collection of interesting ideas) is defined as the minimum of distances between this point and points of the finite set. If an idea-attractor is close to some idea from the interest-database, then the distance between this idea-attractor and the database is also small. If an idea-attractor is far from all interesting ideas, then the distance between this idea-attractor and the database is large. We now define mathematically a measure of interest for an idea-attractor as the following quantity: one over the sum of the distance (between this idea-attractor and the interest-database) and one: [7]

MEASURE OF INTEREST = 1/(DISTANCE +1).

Thus, if the distance is small the measure of interest is large ; if the distance is large it is small.[8]

We now can determine the value of the parameter k (the lowest possible value of the measure of interest). Denote by the symbol L maximum of distances between all possible pairs of mental points. Thus the minimal possible value of the measure of interest is equal to

k=1/(L+1).

Here L can finite as well as infinite. In the latter case k=0. Finally, we remark that, since the minimal distance equals zero, the maximal value of the measure of interest is 1. Thus it takes values in the segment [k, 1].

---

[7] We shift the distance by one to escape the appearance of zero in the denominator.
[8] We can consider the measure of interest as an emotional characteristic (quantative!) of an AI-system. Of course, this general emotion of interest can be essentially refined by introducing a number of data bases corresponding to various forms of interest: love, social communication,…

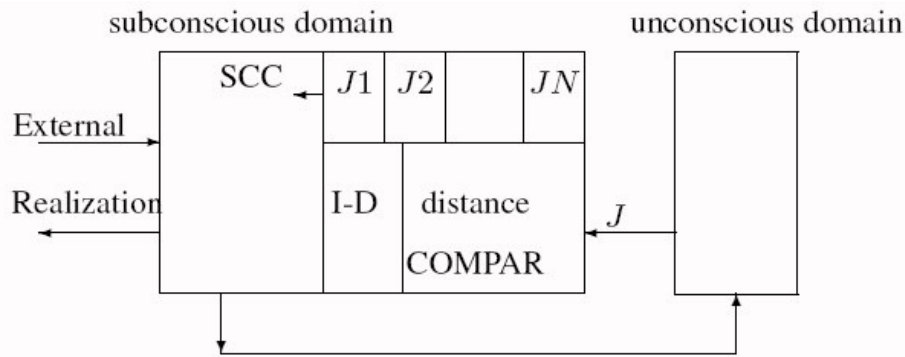

Figure 2: Model 2 of subconscious/unconscious functioning (comparative analysis of ideas)

**The mental architecture of the AI- brain in Model 2 is given by Figure 2:** A new block -- comparator (it is denoted byCOMPAR) -- in the subconscious domain measures the distance between an idea-attractor (it is denoted by J) which has been produced in the unconscious domain and the database of interesting ideas (the latter is denoted I-D. This distance determines the level of interest for an idea-attractor. Ideas waiting for realization (they are denoted J1, J2,..., JN) are collected in a special collector. They are ordered with respect to values of their measures of interest.If for some idea-attractor its measure of interest is larger than the preserving threshold, then the value of interest to this idea does not decrease with time.

### 4.3. Model 3: measure of interdiction

The life of a cognitive system which was described by Model 2 is free of contradictions. Such a cognitive system is always oriented to realizations of the most interesting ideas, wishes, desires. However, environment (and, in particular, social environment) produces some constraints to realizations of some interesting ideas. In a mathematical model we introduce a new quantity which describes a **measure of interdiction** for an idea-attractor. It can be again assumed that the measure of interdiction takes values in the segment [k, 1]. Ideas-attractors with small measures of interdiction have low levels of interdiction. If for some idea its measure of interdiction is approximately equal to k, then this is a "free idea". Ideas with large measures of interdiction have high levels of interdiction. If for some idea its measure of interdiction is approximately equal to one, then such an idea is "totally forbidden." The interdiction function is computed in the same way as the interest function. The subconscious domain contains a **database of forbidden ideas**. [9]Here the notion "forbidden ideas" should be interpreted extremely widely. For example, such a database contains the subject's ethical and other standards which are represented in the form of restrictions.

In Model 3 the comparator measures not only the distance between an idea-attractor (which has been transmitted to the conscious domain from the unconscious domain) and the set of interesting idea, but also the distance between an idea-attractor and the set of forbidden ideas. This distance is defined as minimum with respect to distances between the idea-attractor and all ideas belonging to the interdiction-database. If the attractor is close to some forbidden idea, then the distance is small. If the attractor is far from all forbidden ideas, then the distance is large.

We define the measure of interdiction in the same way as we have defined the measure of interest. This is the following quantity: one over the sum of the distance between this idea-attractor and the interdiction-database and one. The measure of interdiction is large if the distance is small and it is small if the distance is large.

The control center SCC must take into account not only the level of interest of an idea-attractor, but also the level of interdiction of this idea. The struggle between interest and interdiction induces essential features of human psychology.[10] We consider a simple model of such a struggle. For an idea-attractor we define **consistency** (between interest and interdiction) as a linear combination of the measures of interest and interdiction:

---

[9] The measure of interdiction is another emotional characteristic of an AI-system.

$$\text{CONSISTENCY} = a\ \text{INTEREST} + b\ \text{INTERDICTION},$$

where a and b are some real coefficients. Such a linear combination depends on a cognitive system. In the simplest case it can be just the difference between these measures:

$$\text{CONSISTENCY} = \text{INTEREST} - \text{INTERDICTION}$$

Such a functional describes "normal behaviour." A risky person may have e.g. the following functional:

$$\text{CONSISTENCY} = a\ \text{INTEREST} - \text{INTERDICTION},$$

where the coefficient a is sufficiently large. Such a guy would neglect danger and interdiction and he will be extremely stimulated even by a minimal interest. We can consider even an "adrenalin-guy" having

$$\text{CONSISTENCY} = \text{INTEREST} + \text{INTERDICTION}$$

For him danger and interdiction are not less exciting than the interest.

We now modify Model 2 and consider, instead of the realization threshold based on the measure of interest, a realization threshold which is based on the measure of consistency. The presence of such a threshold plays the role of a filter against "inconsistent ideas-attractors." If for some idea-attractor its measure of consistency is larger than the consistency-threshold, then such an idea will stay in the collector of ideas waiting for realization. In the opposite case such an idea will be deleted without any further analysis.

It is convenient to consider a special block in the subconscious domain, an **analyzer**. This block contains: a) the comparator which measures distances from an idea-attractor to databases of interesting and forbidden ideas; b) a computation device which calculates measures of interest, interdiction and consistency; this device also checks consistency of an idea (by comparing it with the realization threshold); c) a transmission device which sends an idea-attractor to the collector or trash. In the model under consideration the order in the queue of ideas in the collector is based on the measure of consistency. It is also convenient to introduce a special block -- **server**, in the subconscious domain which orders ideas in the collector with respect to values of their consistency.

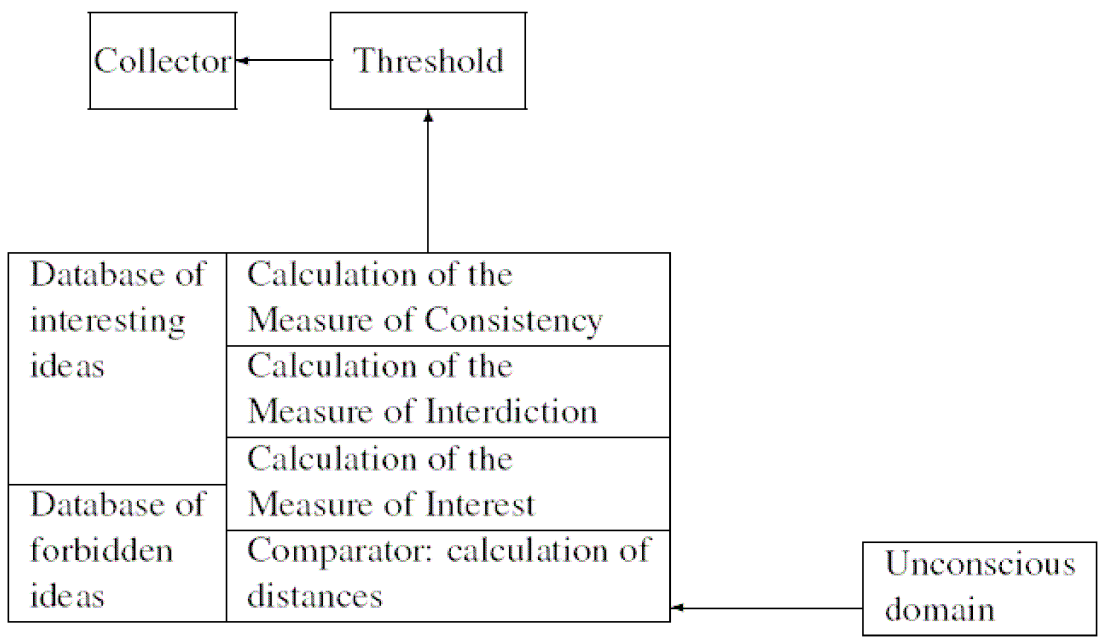

Figure 3: The structure of analyzer

**Comment on Figure 3:** A cognitive system described by Model 3 has complex cognitive behavior. However, this complexity does not imply "mental problems". The use of the consistency functional -- a linear combination of the measures of interest and interdiction -- solves the contradiction between interest and interdiction for an idea-attractor.

We can again assume that there exists a preserving threshold such that ideas-attractors having the consistency larger than this threshold must be realized in any case. It is natural to assume that the preserving threshold is essentially larger than the realization one. This threshold plays the important role in the process of the time evolution of consistency of an idea-attractor in the collector.

We can assume that the consistency-measure decreases exponentially with time (thus this quantity will very quickly become less than the realization threshold and after that this idea will disappear from the collector without any trace and hence it will be never realized). But it will be assumed that if the consistency-measure is larger than the preserving threshold then this measure will not be changed.

The main disadvantage of the cognitive system described by Model 3 is that the analyzer permits the realization of ideas which have at the same time very high levels of interest and interdiction (if the measures of interest and interdiction compensate each other in the consistency function). For example, let the consistency function be equal to the difference between the measure of interest and the measure of interdiction. Assume that the realization threshold is equal to zero. For such a brain the analyzer sends to the collector totally forbidden ideas (with measure of interdiction which is approximately equal to one) having extremely high interest (with measure of interest which is approximately equal to one) Such a behavior (a storm of cravings) can be dangerous, especially in a group of cognitive systems with a social structure. Therefore functioning of the analyzer must be based on a more complex analysis of ideas-attractors which is not reduced to the calculation of the consistency function and comparing it with the realization threshold.

### 4.4. Model 4: forbidden wishes and desires

Suppose that a cognitive system described by Model 3 improves its brain by introducing two new thresholds: the **threshold of maximal interest** and the **threshold of maximal interdiction**. If for some idea-attractor its measure of interest is larger than the maximal interest threshold, then such an idea is extremely interesting. The cognitive system can not simply delete this attractor. If for some idea-attractor its measure of interest is larger than the maximal interdiction threshold, then such an idea is strongly forbidden. The cognitive system can not simply send this idea to the collector to wait for realization. We now introduce the **"domain of doubts"**. These are ideas such that both measures of interest and interdiction are larger that the corresponding maximal thresholds. If an idea-attractor belongs to the domain of doubts, then the cognitive system cannot take automatically (on the basis of the value of the consistency) the decision on realization of this idea.

## 5. Repression

On the one hand, the creation of an additional block in the analyzer to perform analysis of ideas-attractors by comparing them with maximal-interest and maximal-interdiction thresholds plays the positive role. Such a brain does not proceed automatically to realizations of dangerous ideas-attractors, despite their high attraction. We recall that a brain described by Model 3 would proceed totally automatically by comparing the measure of consistency with the realization threshold. On the other hand, this step in the cognitive evolution induces mental problems for a cognitive system. In fact, the appearance of the domain of doubts in the mental space is the origin of some psychical problems and mental diseases. Let the analyser find that an idea-attractor belongs to the domain of doubts— forbidden wish (desire, impulse, experience).[11] The brain is not able neither to realize such an idea nor simply to delete it. What could a brain do in this situation? The answer to this question was given in Freud, 1962a,b: such a forbidden wish is shackled into the unconscious domain.[12]

---

[11] "All these experiences had involved the emergence of a wishful impulse which was in sharp contrast to the subjects other wishes and which proved incompatible with the ethical and aesthetics standards of his personality," Freud, 1962a.

[12] "There had been a short conflict, and the end of this internal struggle was that the idea which had been appeared before consciousness as the vehicle of this irreconcilable wish fell a victim to repression, was pushed out of consciousness with all its attached memories and was forgotten. Thus the incompatibility of the wish in question with the patient's ego was the motive for the repression: the subject's ethical and other standards were the repressing forces. An acceptance of the incompatible wishful impulse or a prolongation of the conflict would have produced a high degree of unpleasure; this unpleasure was avoided by means of repression, which was thus revealed as one of devices serving to protect the mental personality," Freud, 1962b.

In our model, the unconscious domain contains (besides the processing domain and the unconscious control center UC) a special collector for repressed ideas -- **forbidden wishes.** By Freud's it is also a part of Ego (but an unconscious part). After a few attempts to transform an idea-attractor belonging to the domain of doubtful ideas into some non-doubtful idea, SCC sends such a doubtful idea-attractor to the collector for repressed ideas.

What can one say about the further evolution of a doubtful idea in the collector for repressed ideas? It depends on a cognitive system (in particular, a human individual). In principle, this collector might play just the role of a churchyard for doubtful ideas. Such a collector would not have output connections and a doubtful idea
(hidden forbidden desire, wish, impulse, experience) would disappear after some period of time.

## 5.1. Complexes and Symptoms

However, Freud demonstrated that advanced cognitive systems (such as human individuals) could not isolate completely a hidden forbidden wish. They could not perform the complete interment of doubtful ideas in the collector for repressed ideas. In our model, this collector has an output connection with the unconscious control center UC. At this moment the existence of such a connection seems to be just a disadvantage in the mental architecture of a cognitive system. It seems that such a cognitive system was simply not able to develop a neuronal structure for 100\%-isolation of the collector for repressed ideas. However, later we shall see that the cyclic pathway: from SCC to the collector for doubtful ideas, then to UC, and, finally, again to SCC, has important cognitive functions. We might speculate that such a connection was specially created in the process of evolution. But we start with the discussion on negative consequences of existence of this cyclic pathway.

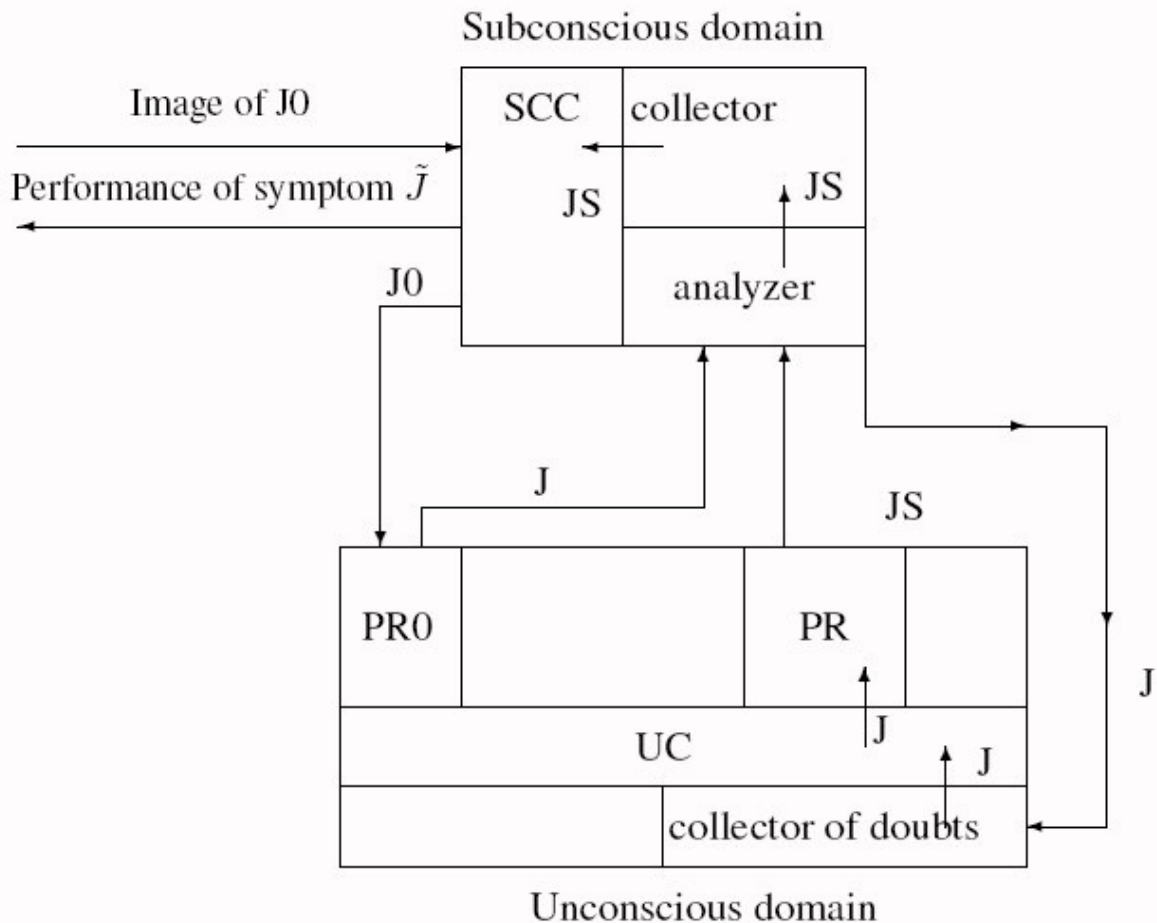

Figure 4: Symptom induced by a hidden forbidden wish

**Comment on Figure 4:** Starting with an initial idea J0 a processor PR0 produces an attractor J; the analyzer computes the measures of interest and interdiction for this idea-attractor; the analyzer considers it as a doubtful idea:
both measures of interest and interdiction are too high, they are larger than the maximal thresholds; SCC sends this idea-attractor to the collector of repressed ideas; it becomes hidden forbidden wish; it moves from the collector of repressed ideas to the unconscious control center UC; UC sends it to some processor PR that produces a new idea-attractor JS. The analyzer may decide that this idea-attractor can be realized (depending on the distances from JS to databases of interesting and forbidden ideas and the magnitude of the realization threshold). In this case the analyzer sends JS through the collector (of ideas waiting for realization) to a performance. Such an idea-attractor JS is a symptom}induced by the original idea-attractor J (in fact, by the initial idea J0).

Our present considerations can be interpreted as creation of AI-models for Freud's theory of subconscious/unconscious mind. In our model an idea belonging to the collector for repressed ideas has the possibility to move to UC. The unconscious control center UC sends this idea to one of the thinking processors. This processor performs iterations starting with this hidden forbidden wish as an initial idea. It produces an idea-attractor. In the simplest case the processor sends its output, an idea-attractor, to the subconscious domain. The subconscious analyzer performs analysis of this idea. If the idea does not belong to the domain of doubts, then the analyzer sends the idea to the collector of ideas waiting for realization. After some period of waiting the idea will be send to realization. By such a realization SCC removes this idea from the collector of ideas waiting for realization. However, SCC does not remove the root of the idea (complex), namely the original hidden forbidden wish, because the latter is now located in the unconscious domain. And SCC is not able to control anything in this domain. A new idea-attractor generated by this forbidden wish is nothing other than its new (unusual) performance. Such unconscious transformations of forbidden wishes were studied in Freud, 1962a,b, 1900. In general a new wish—the final idea-attractor -- has no direct relation to the original forbidden wish. This is nothing but a **symptom** of a cognitive system, cf. Freud, 1962b: "But the repressed wishful impulse continues to exist in the unconscious.} It is on the look-out for an opportunity of being activated, and when that happens it succeeds in sending into consciousness a disguised and unrecognized substitute for what has been repressed, and to this there soon become attached the same feelings of unpleasure which it was hoped had been saved by repression. This substitute for the repressed idea—the symptom—is proof against further attacks of defensive ego; and in place of a short conflict an aliment now appears which is not brought to an end by the passage of time."

## 5.2. Resistance force

A cognitive system wants to prevent a new appearance of forbidden wishes (which were expelled into the collector of repressed ideas) in the subconscious (and then conscious) domain. In our model a brain has an additional analyzer, the unconscious one, (located in the unconscious domain) that must analyze nearness of an idea-attractor produced by some processor and ideas which has been already collected the collector of repressed ideas. The unconscious analyzer contains a comparator that measures the distance between an idea-attractor which has been produced by a thinking block and the database of hidden forbidden wishes: Then this collector calculates the corresponding measure of
interdiction by using the same rule as it was used for the database of forbidden ideas in the subconsciousness. If such an unconscious interdiction is large (approximately one), then this idea-attractor is too close to ne of former hidden forbidden wishes. This idea should not be transmitted to the subconscious (and then conscious) domain.

Each individual has its own **blocking threshold**: if the measure of unconscious interdiction (based on the comparing with the database of hidden forbidden wishes) is less than the blocking threshold, then such an idea-attractor is transmitted into the subconscious and then conscious domains; if this measure is larger than the threshold, then such an idea-attractor is deleted directly in the unconscious domain. In the latter case the idea-attractor will never come to the conscious domain.[13] This blocking threshold determines the degree of blocking of some thinking processors by forbidden wishes. Thresholds can depend on processors. For some individuals (having rather small values of blocking thresholds), a forbidden wish may completely stop the flow of information from some processors to the subconscious domain. The same hidden forbidden wish may play a negligible role for individuals having rather large magnitude of blocking thresholds. Therefore the blocking thresholds are important characteristics which can be used to distinguish normal and abnormal behaviors.[14] In our mental cybernetic model blocking thresholds play the

---

[13] It might happen that analysis in the subconscious domain would demonstrate that this idea has sufficiently high consistency (larger than the realization threshold) and both measures of interest and interdiction are less than corresponding maximum thresholds. In the absence of hidden forbidden wishes this idea-attractor would be realized.

[14] "I found the confirmation that forgotten memories were not lost. They were in the patient's possession and were ready to emerge in association to what still was known by

role of sources of the resistance force which does not permit reappearance of hidden forbidden wishes, desires and wild impulses which were repressed.

We note again that blocking thresholds depends on thinking processors. Thus the same individual can have the normal threshold for one thinking block and abnormal degree of blocking for another thinking block.

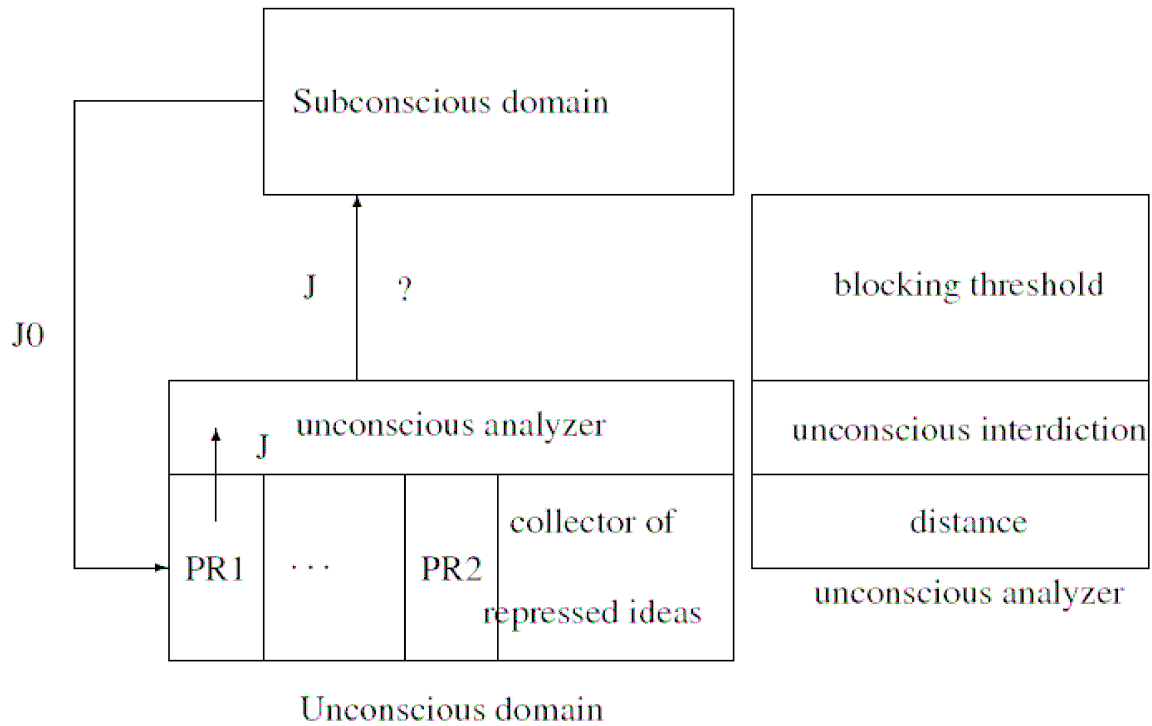

Figure 5: Interference of an idea-attractor with the domain of hidden forbidden wishes. Internal structure of the unconscious analyzer.

**Comment on Figure 5:** The unconscious analyzer computes the distance between an idea-attractor (produced by a thinking block PR1) and the database of hidden forbidden wishes. If this distance is relatively small, i.e., the measure of unconscious interdiction is relatively large, then such an idea-attractor does not go to the subconscious domain.}

## 6. AI-base for the pleasure and reality principles

The pleasure principle is a psychoanalytical term coined by Sigmund Freud. Respectively, the desire for immediate gratification versus the deferral of that gratification. Quite simply, the pleasure principle drives one to seek pleasure and to avoid pain. We shall present AI-justification of this principle. It is convenient to consider the evolution of the pleasure-function through development from Model 1 to Model 4. We start with Model 2. In this model pleasure is identified with the interest-measure. This mental quantity takes its values in the segment $[k, 1]$. Thus we quantified pleasure. The value k corresponds to minimal pleasure and the value 1 to maximal pleasure. This quantity of interest-pleasure is the basis for ordering of ideas attractors for their realization. The brain wants most the ideas having the highest magnitude of pleasure. Such ideas have the highest priority in realization.[15] Moreover, ideas inducing not so much pleasure (e.g., pleasure which little bit extends the value k) might be never realized, because they might just disappear from the collector of ideas waiting for realization. Thus a brain based on Model 2 would like to

---

```
him; but there was some force that prevented them from becoming conscious and compelled
them to remain unconscious. The existence of this force could be assumed with
certainty...", Freud, 1962b
```

[15] The feeling of pleasure is approached at the moment of realization. The strength of this feeling is determined by the magnitude of the interest-measure.

maximize the pleasure-function which is defined on the space of ideas. We recall that the interest-measure increases with the decreasing of the distance from an idea-attractor to the interest-database.

We now consider Model 3. Here a brain can calculate not only the interest-function on the space of ideas, but also the interdiction function. The purpose of the latter one is to prevent such a brain from conflicts with reality. Here we chose a pleasure-reality function is identified with the consistency-function. As was mentioned, the simplest form of the consistency function is simply the difference between the functions of interest and interdiction:

$$\text{PLEASURE—REALITY} = \text{INTEREST} - \text{INTERDICTION},$$

Thus the greatest pleasure and at the same time the greatest consistency with reality is approached in the case of the highest interest and the lowest interdiction, e.g., the interest-measure = 1 and the interdiction-measure = k, so the the pleasure-reality function takes the value 1-k. It is a good point to remark that the pleasure-reality function (given by the measure of consistency) depends on an individual. In general it is an arbitrary linear combination of interest and interdiction:

$$\text{PLEASURE—REALITY} = a\, \text{INTEREST} + b\, \text{INTERDICTION},$$

where a and b some coefficients.

In Model 4 the pleasure-reality function is the same as in Model 3. Finally, we consider Model 1. Here we have only dynamical systems which process external and internal stimuli and produce ideas-attractors which play the role of reactions to those stimuli. Pleasure is approached by realizations of those ideas-attractors. Here Id totally dominates.
We now analyze deeper the structure of the pleasure function. As everything in our models this function is based on the mental distance. Therefore the pleasure principle as well as the reality principle are based on the metric structure of mental space.

## 7. Other approaches to psycho-robots

We do not plan to present here a detailed review on other approaches to psycho-robots. To emphasize differences of our approach from other developments of psycho-robots, we present a citation from the work of Potkonjak el al, 2002: "Man–machine communication had been recognized a long time ago as a significant issue in the implementation of automation. It influences the machine effectiveness through direct costs for operator training and through more or less comfortable working conditions. The solution for the increased effectiveness might be found in user-friendly human–machine interface. In robotics, the question of communication and its user-friendliness is becoming even more significant. It is no longer satisfactory that a communication can be called "human–machine interface", since one must see robots as future collaborators, service workers, and probably personal helpers." In contrast, the main aim of our modeling is not at all creation of friendly helpers to increase their effectiveness. We would like to create AI-systems which would really have essential elements of human psyche. We shown that already psycho-robots with a rather simple AI-psyche – two emotions and two corresponding data bases – would exhibit (if one really wants to simulate human's psyche) very complicated psychological behavior. In particular, they would create various psychical complexes which would be exhibited via symptoms. We also point out to the crucial difference of our "Freudian psycho-robots" from psycho-robots created for different computer game ("psycho-automata").

Our aims are similar of those formulated for humanoid robots, see e.g. Brooks et al., 1981a,b, 1999, 2002. However, we jump directly to high level psyche (without to create e.g. the visual representation of reality). The idea of Luc Steels to create a robot culture via societies of self-educating robots, Manuel, 2003, is also very attractive for us. It is clear that real humanoid psyche (including complexes and symptoms) could be created only in society of interacting Psychots and people. Moreover, such AI-societies of Psychots can be used for modeling psychoanalytic problems and development of new methodologies of treatment of such problems.

## 8. Conclusions

We proposed a series of the AI-type models for advanced psychological behavior. Our approach is based on geometrization of psychological processes via introduction of mental metric space, dynamical processing of mental states and emotional-type decision making based on quantative measures of interest and interdiction and corresponding data bases of ideas. Increasing complexity of AI-modeling implies with necessity appearance of psychological features such as complexes and symptoms which are being handled by psychoanalysis during the last hundred years. Such a complicated behavior has not only negative consequences (e.g. hysteric reactions)[16], but it also plays an important controlling role.

The presented AI-models can be used for creation of AI-systems, which we call psycho-robots (Psychots), exhibiting important elements of human psyche. At the moment domestic robots are merely simple working devices However, in future one can expect demand in systems which be able not only perform simple work tasks, but would have elements of human self-developing psyche. Such AI-psyche could play an important role both in relations between psycho-robots and their owners as well as between psycho-robots. Since the presence of a huge numbers of psycho-complexes is an essential characteristic of human psychology, it would be interesting to model them in the AI-framework. As was already pointed out, complex psychological behavior induces important controlling structures which are self-developing (in the process of interaction of a psycho-robot with human beings or other psycho-robots). However, psycho-robots would pay the same price for complexity of their psyche as it was paid by people. Some psycho-robots would exhibit elements of psychopathic behavior.

One of major contributions of AI and cognitive science to psychology has been the information human thinking in which metaphor brain- as-computer is taken literally. In the present paper we extended the AI-approach to modeling of human psychology. We created the computer-architecture for modeling of very delicate features of human psychological behavior.

---

[16] A hysteric reaction of domestic robot ? Why not?!